# Lightweight Combinational Machine Learning Algorithm for Sorting Canine Torso Radiographs

Masuda A. Tonima
*Department of Mechanical, Industrial and Aerospace Engineering*
Concordia University
Montreal, QC, Canada
masuda.tonima@concordia.ca

Fatemeh Esfahani
*Department of Computer Science*
University of Victoria
Victoria, BC, Canada
esfahani@uvic.ca

Austin DeHart
*R&D Department*
Innotech Medical Industries Corp.
North Vancouver, BC, Canada
austin.dehart@gmail.com

Youmin Zhang
*Department of Mechanical, Industrial and Aerospace Engineering*
Concordia University
Montreal, Canada
youmin.zhang@concordia.ca

*Abstract*—The veterinary field lacks automation in contrast to the tremendous technological advances made in the human medical field. Implementation of machine learning technology can shorten any step of the automation process. This paper explores these core concepts and starts with automation in sorting radiographs for canines by view and anatomy. This is achieved by developing a new lightweight algorithm inspired by AlexNet, Inception, and SqueezeNet. The proposed module proves to be lighter than SqueezeNet while maintaining accuracy higher than that of AlexNet, ResNet, DenseNet, and SqueezeNet.

## I. INTRODUCTION

In today's technologically advanced world, everything is being made easier and enjoyable for human beings; from brushing teeth to operating on the brain, everything is technology-aided and made more comfortable. Among these advancements, human's best friends, animals, have not been left far behind; cameras, trackers, collars, fences, toys, etc., have been technologically altered to accommodate these fantastic beasts [1]. In the past decade, many technologies solely used on humans are now being adapted for animals. Examples of such technologies are wearable devices, 3D printing used in treatment: Magnetic resonance imaging, ultrasound, x-ray, etc., have become the standard for animal diagnoses [2]. In 2018, The North American Pet Health Insurance Association (NAPHIA) reported a rise of $0.27 billion in pet health insurance premiums to 1.42 billion dollars a year solely in North America, among which canine health represents approximately 90% of this figure [3]. As pet owners keep investing in their pets' wellbeing (~$1400 annually) [3], the demand for better healthcare for these said pets is on the rise. About 90% of Americans consider their pets as part of their family; thus, it is only logical that the same technology used for human diagnosis be applied to these canines [4]. The ultimate goal for technology in diagnostic medicine must be automation, as speed and precision will always be endeavored. Automation also leads to treatments becoming more affordable, safer, uniform, and can carry less risk as equipment longevity improves and human interference is reduced [5]. Burti *et al*. [6] employed machine learning to detect heart size abnormalities from plain radiographs, making it clear that diseases can be autonomously detected by implementing artificial intelligence (AI). The application of this concept requires several steps such as *preprocessing*, *segmentation*, *feature extractions*, etc. The problem of implementing these in veterinary science is that there are vast numbers of inconsistent data that need sorting before the processes can even start to be employed.

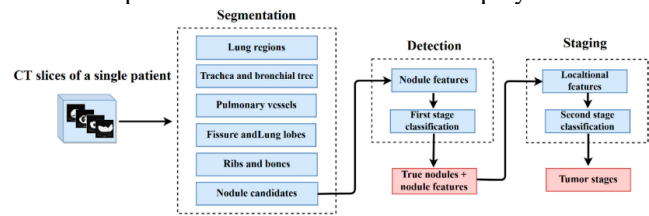

Figure 1 Typical architecture in diagnosis automation [7]

Hence, although Paing *et al*. [7] suggests segmentation is the first step of automation in diagnosis procedure, as shown in Figure *1*, it is suggested that added steps of preprocessing and sorting are required before this (for animals) as capturing stationary radiographs becomes difficult without sedation. This paper explores preprocessing, which sorts mages based on image perspective (dorsal or lateral) and image interest region (upper or lower) in the torso. This step's results will become more useful when a massive amount of data is processed for training algorithms. In this paper, a new convolutional neural network (CNN) architecture is developed, which incorporates different modules to keep the model deep enough while maintaining the number of parameters in the model to be only 0.1 million. Experimental results confirm the efficiency and accuracy of the proposed model for x-ray image classification.

## II. METHODOLOGY

### A. *X-ray imaging for canines*

An essential part of automation in diagnosis via image is the dataset. In this section, radiographs of canine torso are explored. Capturing radiographic images of animals such as cats or dogs can be problematic as movement possesses a huge issue; hence it is crucial that "shutter speed" for these exposures be less than 1/30th of a second [8]. Proper positioning and collimation are of significant importance for creating repeatable images, especially of the torso [9]. There are two views radiographers require to make proper use of radiographs (whether it is in digital or film format): 1) Lateral (LAT) and 2) Ventrodorsal (VD/DV).

For lateral views of thoracic and abdominal radiographs (upper and lower torso regions, respectively), a canine is placed on the table with one side down while the collimated area is marked. To get a proper and steady image, the thoracic limbs must be taped together and pulled away from the area of interest. Resulting in an image of the region of interest that is not superimposed with that of the limb muscle (as to not obscure the view of the desired anatomy) [9]. This is shown in Figure *2*. Additional aids such as a wedge or tape for moving pelvic limbs away can also be used, as shown in Figure *3*. The sternum (breastbone) of the dog must be present in a thoracic radiograph as this ensures the inclusion of vital organs that may be of interest; however, in larger dogs, there might be a need to take separate images to include both the sternum and the lower spinal columns [9].

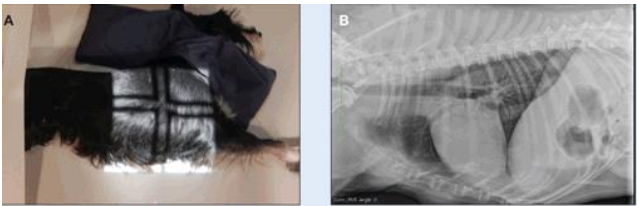

Figure 2 (A) Dog with taped thoracic limbs pulled away from the region of interest for the radiograph. (B) Lateral thoracic radiograph without superimposition of paws due to pulled away thoracic limbs showed in (A) [9]

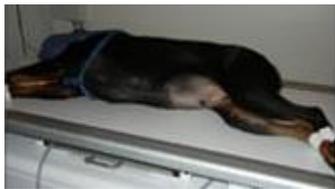

Figure 3 Large dog positioned laterally [8]

Ventrodorsal and Dorsoventral images can be hard to perform as canine positioning for this type of radiograph is difficult. For the former, the dog may have its limbs taped and pulled away from the torso, and with the additional help of a muzzle, a steady radiograph capture is possible. However, it is very cumbersome to make a dog patiently sit in a frog-like position long enough to take a clear image of the latter. How these images are taken is shown in Figure *4* and Figure *5*.

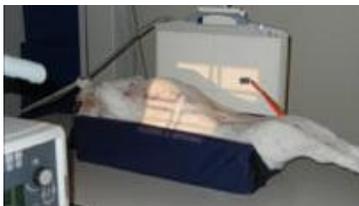

Figure 4 Ventrodorsal image positioning [9]

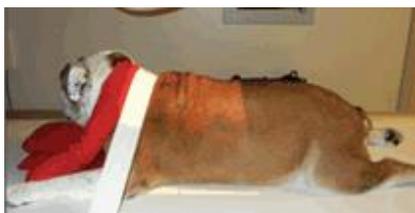

Figure 5 Dorsoventral image positioning [9]

Radiographs are usually the first line of non-invasive tests performed. Correctly positioned high-quality images improve assessment accuracy and can render invasive test methods unnecessary in certain cases. Broken bones, arthritis, and foreign items swallowed by dogs are only a few examples of the many conditions which can be identified through radiography.

*B. Algorithm inspiration*

Despite being first proposed in 1959 by Hubel and Wiesel [10], the convolutional neural network did not gain popularity until later in the century due to its taxing computational needs until LeCun et al. proposed LeNet in 1989 [11] and finally realizing it in 1998 [12]. LeNet has become the father of all modern CNN architectures. It is the first to have utilized stacking convolutions with activation functions, pooling layers, and summing with dense layers in the end [13], leading to a usable CNN that does not exhaust computational capabilities, thus standardizing the method. Since then, many CNN architectures have been established; the family tree of algorithms derived from LeNet is given in Figure *6*, where each derivation brings a significant update such as: AlexNet, which replaces the activation function of LeNet (hyperbolic tangent) with ReLU and uses max-pooling instead of average pooling, leading to reduced computation time; and VGG16 that goes twice as deep as its predecessor by using uniform convolutions leading to more accurate classifications [13].

With each update, the idea is to maintain performance integrity while decreasing parameters and reducing training time without sacrificing depth. The Inception algorithm capitalizes on this idea and stacks blocks of convolutional layers; the latter-versions improve on this idea. The Xception algorithm takes it to another level, implementing depth-wise separable convolutions. Resnet, on the other hand, branches off and delves into skip connection and batch normalizations, ResNext, as the name suggests, takes this further and implements parallel towers within a module. Combining the two branches, parallel towers idea and inception block concept, the Inception-Resnets manage to maintain higher prediction accuracy while training for fewer epochs [14]. While being accurate and useful, these algorithms fail to function with limited memory availability, giving way to algorithms such as SqueezeNet and MobileNet. These algorithms utilize smaller deep neural network (DNN) and depth-wise separable convolutions, respectively, to build lighter systems that may function for FPGA, mobile, and embedded systems [15] [16].

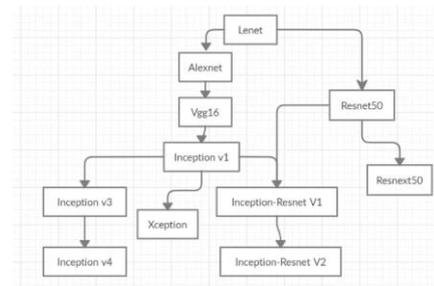

Figure 6 Algorithms derived from LeNet

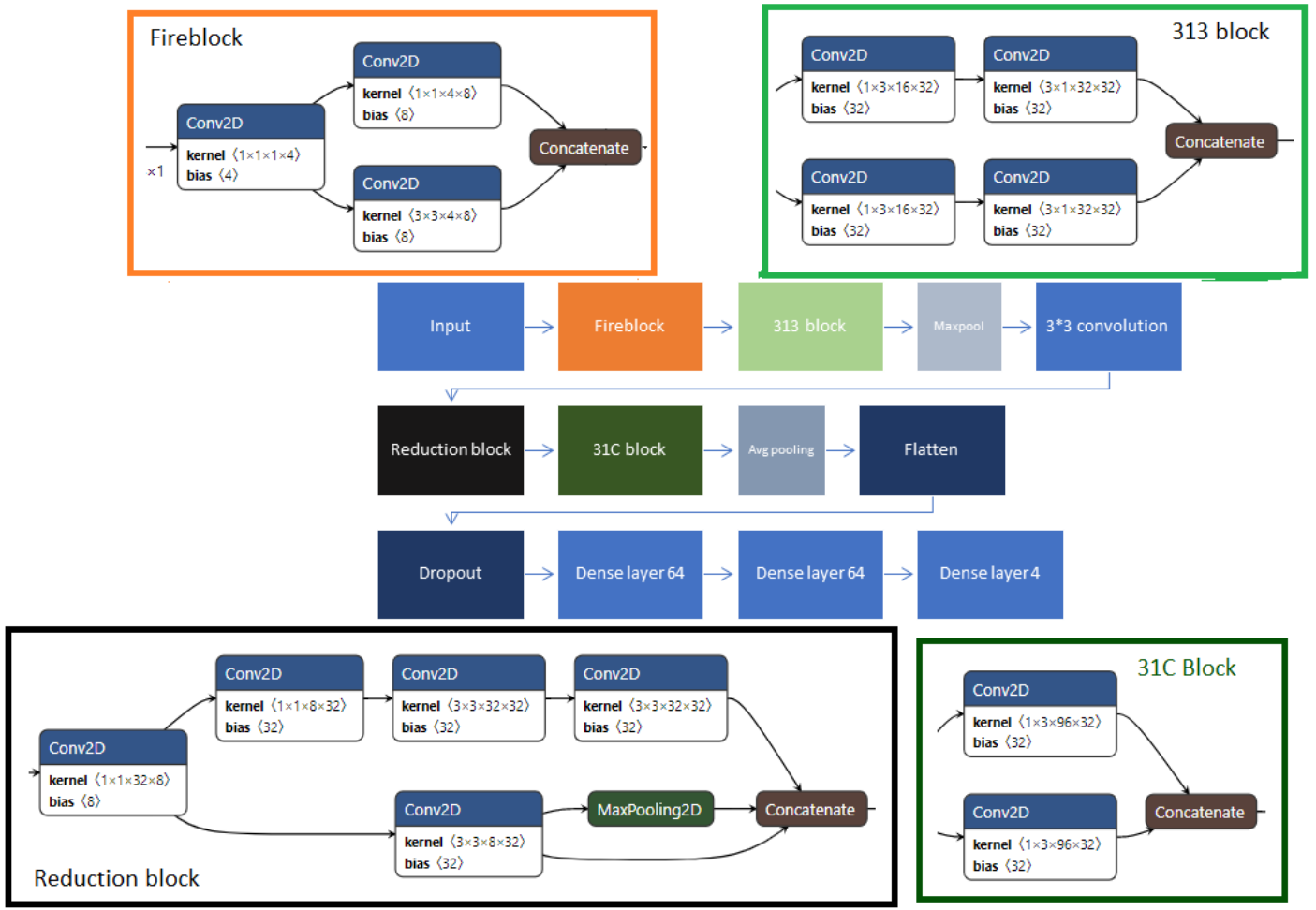

Figure 7 Neural network architecture [17]

## C. Algorithm

This paper investigates all aspects of the above-mentioned algorithms that positively affect the classification of radiographs based on subject and view, and implement concepts such as the Squeeze-Expand module, Reduction block, and the Fire-module. Thus, this algorithm is both CNN and DNN inspired.

Figure 7, shows the architecture block diagram of the algorithm, it shows how the architecture employs various ideas, such as:

- The Fire block: it is a fire module inspired block that accommodates the reduction of parameters from the get-go. It implements a Squeeze followed by two Expand layers, the output of which is then concatenated.
- The 313-block: An Inception-inspired block that helps in maintaining the integrity of prediction accuracy while having reduced computations by employing a set of 1×3 followed by 3×1 kernels in parallel before concatenation.
- The 31C block: another Inception inspired block that helps in maintaining the integrity of prediction accuracy while having reduced computations by employing two parallel 1×3 convolutional blocks followed by concatenation.
- The reduction block: contains a 1×1 convolution that branches its results to a set of 1×1 and two 3×3 convolutions and 3×3 convolutions—Max-pooling and concatenating the result before producing the blocks final output.

The output of the Fire-block is passed through the 313-block. Which is then concatenated, followed by max-pooling with a stride of 4, and then a convolutional layer of 3×3 kernel. This is then passed through the reduction layer and finally the 31C layer. The architecture ends with a dropout of 20% to ensure the reduction of overfitting. The dense layers are automatically tuned depending on the data type, using activation ReLU for the first two and softmax for the last layer. The model summary details are given in [18].

## III. RESULTS AND COMPARISON

In this section, the experimental results are shown in two different settings: 1) the effect of two variations of activation functions, ReLU and Swish, are investigated when used for the convolutional layers in the model; 2) the comparison between the performance of the proposed model and different models mentioned in the literature. Table 2 shows the performance of the model with ReLU and Swish activation functions in terms of precision, recall, F1-score, and accuracy metrics, as well as training time per epoch.

The swish activation function is a combinational function, linear and sigmoid combined, that is a soft-monotonic function bounded below and unbounded above [19].

The swish version of the model better predicts the upper torso images from the lateral view, whereas the rest of the attributes are better predicted by the ReLU version. Overall, the latter works better with the number of images presented.

A comparative study for AlexNet, ResNet50, and DenseNet with the model created is performed. We use a shallow version of DenseNet with compression rate $\theta = 0.5$, $k = 24$, and repetitions [4,4,4,4]. The results are shown in Table 3, and the best performance in each category is underlined. The metrics considered are the same as before. The model created here outperforms or performs just as well as the rest in 9 categories. This includes lower parameter numbers and low training time, which is extremely desirable in medical image processing.

For instance, the average F1-score obtained by the proposed model is about 82%, while DenseNet classifies the images from different classes with a 75% average F1-score only. Comparing the model's performance with AlexNet shows that F1-scores for most classes are higher than the ones obtained by AlexNet, which is quite impressive for a lightweight model with about 195 times fewer parameters than AlexNet. Additionally, the model obtains higher values of precision, recall, and F1-score metrics for DV Lower and LAT Lower classes, while completing training time in only about 19 seconds per epoch compared to ResNet50, which takes 34 seconds. In general, the proposed architecture's performance is quite good when applied to this limited dataset while being lightweight and efficient.

Table 1 Neural network model summary [18]

| Layer (type) | | Kernel size | Activation | Output Shape | Parameters | Connected to |
|---|---|---|---|---|---|---|
| Input | | | | (224, 224, 1) | 0 | |
| Fire block | Squeeze (Conv2D) | 1*1 | Swish/ReLU | (224, 224, 4) | 8 | input |
| | Expand1 (Conv2D) | 1*1 | Swish/ReLU | (224, 224, 8) | 40 | Squeeze |
| | Expand2 (Conv2D) | 3*3 | Swish/ReLU | (224, 224, 8) | 296 | Squeeze |
| Concatenate 1 | | Axis=3 | | (224, 224, 16) | 0 | Expand1 & Expand2 |
| 313 block | C1(Conv2D) | 1*3 | Swish/ReLU | (224, 224, 32) | 1568 | Concatenate 1 |
| | C2(Conv2D) | 1*3 | Swish/ReLU | (224, 224, 32) | 1568 | Concatenate 1 |
| | C11(Conv2D) | 3*1 | Swish/ReLU | (224, 224, 32) | 3104 | C1 |
| | C21(Conv2D) | 3*1 | Swish/ReLU | (224, 224, 32) | 3104 | C2 |
| Concatenate 2 | | Axis=3 | | (224, 224, 64) | 0 | C11 & C21 |
| Max pooling 1 | | Pool size = (4,4) | | (56, 56, 64) | 0 | Concatenate 2 |
| Convolution (Conv2D) | | 3*3 | | (56, 56, 32) | 18464 | Max pooling 1 |
| Reduction block | C (Conv2D) | 1*1 | Swish/ReLU | (56, 56, 8) | 264 | Convolution |
| | C1(Conv2D) | 1*1 | Swish/ReLU | (56, 56, 32) | 288 | C |
| | C2(Conv2D) | 3*3 | Swish/ReLU | (56, 56, 32) | 2336 | C |
| | C11(Conv2D) | 3*3 | Swish/ReLU | (56, 56, 32) | 9248 | C1 |
| | C12(Conv2D) | 3*3 | Swish/ReLU | (56, 56, 32) | 9248 | C11 |
| | Max pooling 2 | Pool size= (1,1) | | (56, 56, 32) | 0 | C2 |
| Concatenate 3 | | Axis=3 | | (56, 56, 96) | 0 | Max pooling 2 & C2 &C12 |
| 31C block | Max pooling 3 | Pool size= (4,4) | | (14, 14, 96) | 0 | Concatenate 3 |
| | C1 (Conv2D) | 1*3 | Swish/ReLU | (14,14, 32) | 9248 | Max pooling 3 |
| | C2 (Conv2D) | 1*3 | Swish/ReLU | (14,14, 32) | 9248 | Max pooling 3 |
| | Concatenate 4 | Axis=3 | | (14, 14, 64) | 0 | C1 & C2 |
| Average pooling | | Pool size= (2,2), stride=4 | | (4, 4, 64) | 0 | Concatenate 4 |
| Flatten | | | | 1024 | 0 | Average pooling |
| Dropout | | Rate=0.2 | | 1024 | 0 | Flatten |
| Dense 1 | | | ReLU | (None, 64) | 65600 | Dropout |
| Dense 2 | | | ReLU | (None, 64) | 4160 | Dense 1 |
| Dense 3 | | | SoftMax | (None, 4) | 260 | Dense 2 |
| Total params: 138,052 Trainable params: 138,052 Non-trainable params: 0 | | | | | | |

Table 2 Variation in results based on activation function.

| | | ReLU | Swish |
|---|---|---|---|
| Precision | DV Upper | **0.85** | 0.75 |
| | DV Lower | **0.94** | 0.90 |
| | LAT Upper | 0.50 | **0.74** |
| | LAT Lower | **0.92** | 0.85 |
| Recall | DV Upper | **0.77** | 0.71 |
| | DV Lower | 0.96 | **0.97** |
| | LAT Upper | **0.55** | 0.54 |
| | LAT Lower | 0.90 | **0.91** |
| F1-score | DV Upper | **0.81** | 0.73 |
| | DV Lower | **0.95** | 0.93 |
| | LAT Upper | 0.52 | **0.63** |
| | LAT Lower | **0.91** | 0.88 |
| Accuracy | | 90 | 84 |
| Training accuracy | | 98% | 95% |
| Validation accuracy | | 90% | 86.2% |
| Parameters | | 0.1M | 0.1M |
| Training time/epoch (s) | | 16 | 17 |

Table 3 Comparison of results of the developed model with AlexNet, ResNet50, and DenseNet

| | | AlexNet | ResNet50, ImageNet weight | DenseNet | ReLU/Swish |
|---|---|---|---|---|---|
| Precision | DV Upper | 0.85 | 0.94 | 0.76 | **0.85** |
|  | DV Lower | 0.77 | 0.93 | 0.73 | **0.94** |
|  | LAT Upper | 0.89 | 0.89 | 0.88 | 0.74 |
|  | LAT Lower | 0.85 | 0.91 | 0.68 | **0.92** |
| Recall | DV Upper | 0.91 | 0.97 | 0.88 | 0.77 |
|  | DV Lower | 0.80 | 0.93 | 0.73 | **0.97** |
|  | LAT Upper | 0.86 | 0.94 | 0.64 | 0.55 |
|  | LAT Lower | 0.80 | 0.83 | 0.77 | **0.91** |
| F1-score | DV Upper | 0.88 | 0.95 | 0.81 | 0.81 |
|  | DV Lower | 0.79 | 0.93 | 0.73 | **0.95** |
|  | LAT Upper | 0.82 | 0.92 | 0.74 | 0.63 |
|  | LAT Lower | 0.87 | 0.87 | 0.72 | **0.91** |
| Accuracy |  | 0.84 | 0.92 | 0.75 | **90** |
| Train accuracy |  | 99.7% | 98.7% | 90% | **98%** |
| Val accuracy |  | 90% | 87% | 81% | **90%** |
| parameters |  | 19.53M | 23.6M | 0.56M | **0.1M** |
| Training time/epoch(s) |  | 27.53s | 34.09 | 27.42 | **19** |

The image classification examples from both versions of the algorithm are shown in **Error! Not a valid bookmark self-reference.**. Row 1 shows images correctly classified by the swish version of the algorithm; from which it is evident that the model can distinguish and sort between DV image subclasses even though the images are similar.

Table 4 Image classification examples, correctly done for first 2 rows

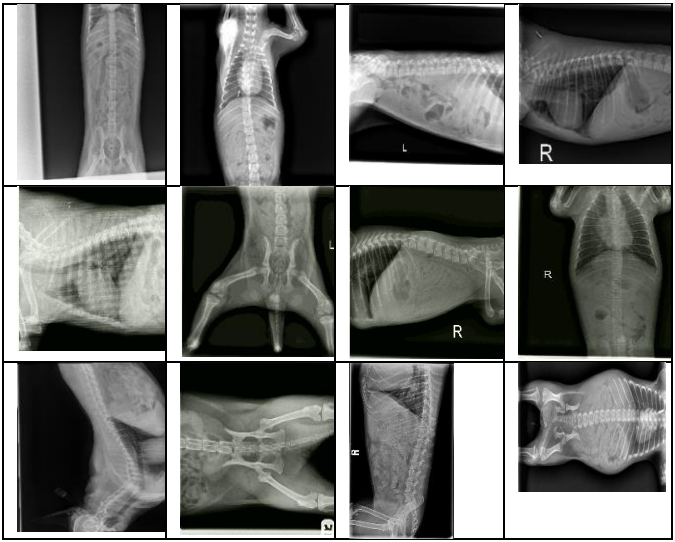

Table 5 Performance Comparison with SqueezeNet

|  |  | SqueezeNet | ReLU/Swish |
|---|---|---|---|
| Precision | DV Upper | 0.65 | **0.85** |
|  | DV Lower | 0.62 | **0.94** |
|  | LAT Upper | 0.67 | **0.74** |
|  | LAT Lower | 0.68 | **0.92** |
| Recall | DV Upper | 0.81 | 0.77 |
|  | DV Lower | 0.60 | **0.97** |
|  | LAT Upper | 0.75 | 0.55 |
|  | LAT Lower | 0.46 | **0.91** |
| F1-score | DV Upper | 0.72 | **0.81** |
|  | DV Lower | 0.61 | **0.95** |
|  | LAT Upper | 0.71 | 0.63 |
|  | LAT Lower | 0.54 | **0.91** |
| Accuracy |  | 0.65 | **90** |
| Train accuracy |  | 92% | **98%** |
| Validation accuracy |  | 75% | **90%** |
| parameters |  | 0.73M | **0.1M** |
| Runtime/epoch(s) |  | 27.73 | **16** |

The only difference being that the heart is visible in "DV-upper" images and the pelvis in the "DV-lower". This is the kind of classification one expects from a human, and the swish-model performs well enough to distinguish between the two. Similarly, despite having similar central positioning for the lateral images, the swish-model can differentiate between the upper and lower torso images where, in one, the heart is visible, and in the other it is not. The example images were purposefully chosen to be similar to show the distinguishing capability of the model. Similarly, row 2 displays the results of the ReLU version of the algorithm, which correctly identifies DV/VD/Lateral and Upper/Lower images accurately despite subclasses having close resemblance. However, both versions fail to classify correctly when images are rotated, as shown in the 3$^{rd}$ row. This model is further compared with SqueezeNet's performance as it is known for its minimal memory requirement, shown in Table *5*.

Even when compared with SqueezeNet, the number of parameters of the model developed here is lower, i.e., smaller, and lightweight. The overall performance is better in comparison with SqueezeNet.

## IV. CONCLUSION

This study is the first step of many in the process of developing diagnostic automation with the use of canine radiographs. The authors' deep neural network model successfully sorts this type of data, view-wise and location-wise, even when compared with established algorithms such as AlexNet and proves to be smaller in size than SqueezeNet. For these reasons, implementing this architecture in small and portable hardware has become more feasible. The next step in this study is the detection of specific anatomy, which will further allow the detection of normality or abnormality in the radiographs sorted through this algorithm.

## V. ACKNOWLEDGMENTS

The work reported in this paper was financially supported in part by a MITACS Accelerate Project (no. FR56849) under the partner organization Innotech Medical Industries Corp. And the Natural Sciences and Engineering Research Council of Canada (NSERC).